\documentclass[letterpaper, 10 pt, conference]{ieeeconf}  

\IEEEoverridecommandlockouts                              

\usepackage{authblk}
\usepackage[ruled,vlined]{algorithm2e}
\usepackage{multicol}
\usepackage{inputenc}
\usepackage{amsfonts}
\usepackage{amsmath}
\usepackage{amssymb}
\usepackage[T1]{fontenc}
\usepackage{microtype}
\usepackage[dvipsnames]{xcolor}
\usepackage{graphicx}
\usepackage{graphics}
\let\labelindent\relax
\usepackage{enumitem}
\usepackage[font=footnotesize]{caption}
\usepackage{subcaption}
\usepackage{booktabs}
\usepackage{multirow}
\usepackage[flushleft]{threeparttable}
\usepackage[export]{adjustbox}
\usepackage{breqn}
\usepackage{acronym}
\usepackage{setspace}
\usepackage{bm}
\usepackage{stackengine}
\usepackage{siunitx}
\usepackage{makecell}
\usepackage{smartdiagram}
\usepackage{tikz}
\usesmartdiagramlibrary{additions}
\usepackage{hyperref}
\definecolor{azure}     {rgb}{0,0.5,1}
\definecolor{dkpowder}  {rgb}{0,0.2,0.7}
\definecolor{deepred}   {rgb}{0.7,0,0}
\definecolor{deepblue}  {rgb}{0,0,0.7}
\definecolor{deepgreen} {rgb}{0,0.5,0}
\definecolor{deeporange}{rgb}{0.91, 0.41, 0.17}
\hypersetup{%
	pdfpagemode  = {UseOutlines},
	bookmarksopen,
	pdfstartview = {FitH},
	colorlinks,
	linkcolor = {dkpowder},
	citecolor = {dkpowder},
	urlcolor  = {dkpowder},
}
\usepackage{listings}
\renewcommand{\thetable}{\arabic{table}}

\graphicspath{{/images}}
\makeatletter
\def\endthebibliography{%
	\def\@noitemerr{\@latex@warning{Empty `thebibliography' environment}}%
	\endlist}
\makeatother
\definecolor{label-running} {RGB}{ 31,119,180}
\definecolor{label-walking} {RGB}{255,127, 14}
\definecolor{label-jumping} {RGB}{ 44,160, 44}
\definecolor{label-standing}{RGB}{148,103,189}
\definecolor{label-sitting} {RGB}{140, 86, 75}
\definecolor{label-lying}   {RGB}{127,127,127}
\definecolor{label-falling} {RGB}{188,189, 34}
\definecolor{label-transit} {RGB}{ 23,190,207}

\title{\LARGE \bf
A Multi-Simulation Approach with Model Predictive Control\\ for Anafi Drones
}

\author{Pascal Goldschmid$^{1}$ and Aamir Ahmad$^{1,2}$
\thanks{The authors thank Curtis Walch and Christian Gall for their support with the flight experiments. The authors thank the International Max Planck Research School for Intelligent Systems (IMPRS-IS) for supporting Pascal Goldschmid.}
\thanks{$^{1}$University of Stuttgart, Faculty of Aerospace Engineering and Geodesy, Institute of Flight Mechanics and Control (iFR), Flight Robotics and Perception Group (FRPG). Pfaffenwaldring 27, 70569 Stuttgart, Germany
        {\tt\small \{pascal.goldschmid, aamir.ahmad\}@ifr.uni-stuttgart.de}}%
\thanks{$^{2}$Max Planck Institute for Intelligent Systems, T\"ubingen, Perceiving Systems Department. Max-Planck-Ring 4, 71069 T\"ubingen, Germany
     }%
}
\begin{document}

\maketitle
\thispagestyle{empty}
\pagestyle{empty}

\begin{abstract}
Simulation frameworks are essential for the safe development of robotic applications. However, different components of a robotic system are often best simulated in different environments, making full integration challenging. This is particularly true for partially-open or closed-source simulators, which commonly suffer from two limitations: (i) lack of runtime control over scene actors via interfaces like ROS, and (ii) restricted access to real-time state data (e.g., pose, velocity) of scene objects.
In the first part of this work, we address these issues by integrating aerial drones simulated in Parrot’s Sphinx environment (used for Anafi drones) into the Gazebo simulator. Our approach uses a mirrored drone instance embedded within Gazebo environments to bridge the two simulators. 
One key application is aerial target tracking, a common task in multi-robot systems. However, Parrot’s default PID-based controller lacks the agility needed for tracking fast-moving targets. To overcome this, in the second part of this work we develop a model predictive controller (MPC) that leverages cumulative error states to improve tracking accuracy. Our MPC significantly outperforms the built-in PID controller in dynamic scenarios, increasing the effectiveness of the overall system.
We validate our integrated framework by incorporating the Anafi drone into an existing Gazebo-based airship simulation and rigorously test the MPC against a custom PID baseline in both simulated and real-world experiments.

Source code is available at: \url{https://github.com/robot-perception-group/anafi_sim}

\end{abstract}

\section{Introduction}\label{sec:Introduction}
Simulators play a fundamental role in robotics, e.g.  for developing dynamic multi-robot environments, control algorithms or action sequences.
Different simulators exist for different tasks, however, over the last years the open-sourced Robot Operating System (ROS) in combination with the physics simulator and visualization tool Gazebo \cite{quigley09}  has emerged as a standard environment  \cite{sarabakha2023}. \\
\begin{figure}[thpb]
      \centering
		\includegraphics[scale=0.25]{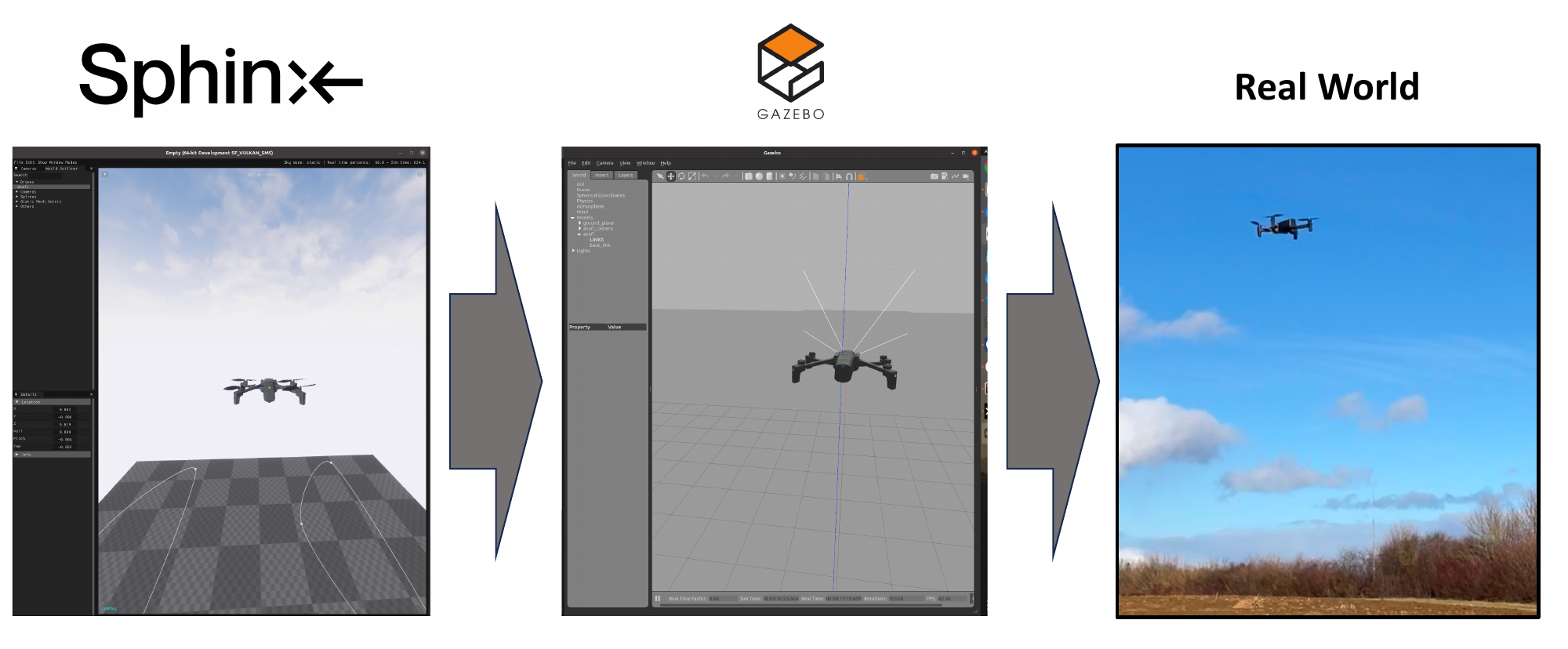}
\caption{Drones simulated in Sphinx (left image) are mirrored in Gazebo (central image), where they can be included in other, multi-robot simulations. Furthermore, our developed control frameworks allow accurate tracking of moving targets in simulation and the real world (right image) a common use case for which our simulation framework is a suitable choice.}
\label{fig:title_image}
 \end{figure}
It is common in research to design and build customized robots for specific purposes and experiments, which is usually a time-consuming process until all hardware-related issues have been solved \cite{pedre2014}. In this regard, commercial robotic platforms often provide a shortcut to quick deployment of a robotic system since these issues have been ruled-out by the manufacturer during the development phase of the product.  However, for scientific robotic applications this advantage is often diminished when no or only a simulation with limited interfaces or customization options is provided along with the robot. 

This problem is aggravated in complex multi-robot dynamic scenarios when robots of different types and different manufacturers need to be simulated at once. 
There are two ways to solve this problem.  First, either all robotic platforms are realized in a common simulation environment or second, the different simulations are interfaced and thus capable of exchanging data. Whereas the first way is often associated with a time consuming system identification with subsequent (re-)implementation of the real robots in the chosen simulator, the second way promises a quicker way to deployment of the system and therefore is the approach we follow in this paper. However, in this regard, often proprietary partially-open or closed-sourced simulators  suffer from two main accessibility problems that limit their utility to the scientific community. i) The state of dynamic actors in the simulation environment cannot be updated via common middleware such as ROS and ii) reading state values such as pose, velocity, accelerations etc. during runtime is blocked.

 In aerial robotics, the drones provided by Parrot Drone are known for having long flight times and flight range and are equipped with powerful hardware and sensors for different applications as illustrated in \cite{sarabakha2023}, thus making them attractive also for scientific purposes. The publicly available simulator Sphinx \cite{parrot_sphinx}  allows the simulation of the drones \textit{Anafi} \cite{parrot_anafi} and \textit{Anafi Ai} \cite{parrot_anafi_ai} in customizable environments. It is also used by Parrot's engineers for development and can therefore be considered as a high-fidelity simulation environment for the  Anafi drone series. However, even-though scenes in which the drone operates can be customized using static and dynamic actors, being only partially open-source Sphinx suffers from the two aforementioned accessibility problems. This is a major drawback since the information about objects in the scene other than the Anafi drone is frequently required for scientific purposes, e.g. during the design of a tracking controller or in multi-robot simulations. This  therefore constitutes the first limitation in using Sphinx we address in this work.

\begin{figure}[thpb]
      \centering
		\includegraphics[scale=0.37]{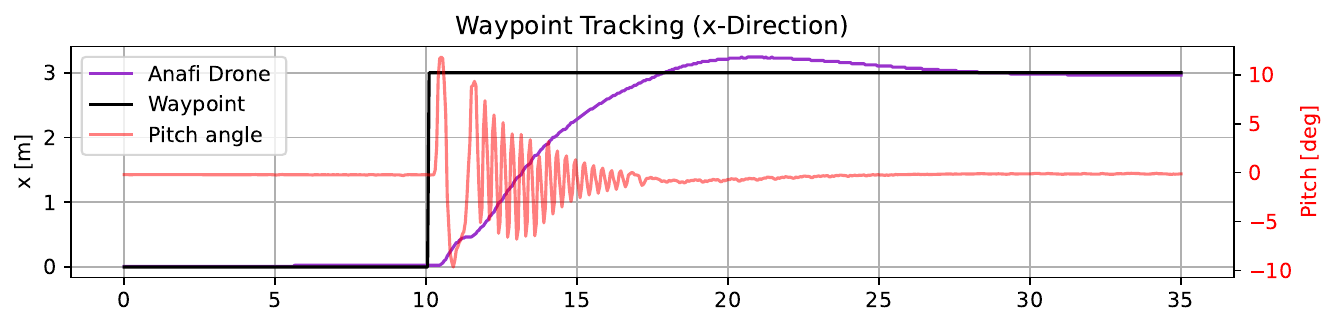}
\caption{Static waypoint tracked with the default PID-based controller (x-direction displayed only). The pitch angle reveals the oscillating motion and slow response. }
\label{fig:moveto}
 \end{figure}

The second limitation of Sphinx we solve in this work is the PID-based controller framework provided with the firmware of the Anafi drone. A common application in aerial robotics is target tracking, e.g. \cite{tallamraju2018,tallamraju2019}. However, depending on the manoeuvrability of the target, agile tracking controllers are required. Feedback controllers such as the provided PID-based control framework are suitable only for slowly moving targets as otherwise large tracking offsets can occur. Furthermore, the PID control stack integrated in the firmware of the Anafi drone shows a peculiar behaviour considerably limiting its applicability to target tracking tasks. Frequent updates to the target waypoint, a common practice in tracking scenarios, do not yield the anticipated smooth transition between successive waypoints of the drone. Instead, an oscillating motion characteristic can be observed, leading to slow response  behaviour which is illustrated by Figure \ref{fig:moveto}.

A starting point to create a solution is given by the attitude controller of the Anafi drone which accepts a continuous stream of roll, pitch, yaw and vertical velocity commands \cite{olympe_pcmd}. \\
Against this background, we address the aforementioned accessibility issues of Sphinx as well as the peculiar behaviour of the PID-based controller framework by introducing the following novel methods, as is illustrated by Figure \ref{fig:title_image}.
\begin{enumerate}
\item \textit{Sphinx-Gazebo integration}: We propose a ROS package that runs Sphinx in parallel with Gazebo. Being the better established framework, Gazebo is expected to contain the majority of simulated robots into which the Anafi drone shall be integrated. We achieve this by creating a visualization of a  drone in Gazebo, which mirrors the flight behaviour of the Anafi drone simulated in Sphinx. Communication between the simulators is realized using ROS and the \textit{anafi_ros} package \cite{sarabakha2023}. 
\item \textit{Target tracking control}: Using a novel concept of cumulative error states, we implement a quadratic MPC building on top of  \cite{tallamraju2018}. It is specifically  designed to reduce the target tracking error and runs efficiently with low computational cost. We also implement a customized PID-based controller in ROS solving the unwanted oscillating motion characteristic present in the  PID-framework provided in the \textit{Anafi} firmware. To this aim, the controller framework's output are roll, pitch, yaw and vertical velocity commands, which can be sent to the drone in a continuous fashion. We use our PID-based controller framework to compare the performance of our MPC against it.
\end{enumerate}
We demonstrate the efficacy of our two solution approaches as follows. First, we include the Anafi drone in the airship simulation \cite{price2022}.  Second, we evaluate the performance of the MPC in simulated and real-world experiments by comparing it to our customized PID controller baseline.
The remainder of this paper is structured as follows. Related work is summarized in Section \ref{sec:related_work}, Section \ref{sec:methodology} contains information about the connection between Sphinx and Gazebo as well as the controller designs, followed by Section \ref{sec:implementation} explaining details of the implementation. The results are described in Section \ref{sec:experiments} before we conclude this work in Section \ref{sec:conclusion}.

\section{Related Work} \label{sec:related_work}
In general, there exist various universal purpose robotic simulators, e.g. Pybullet \cite{pybullet}, NVIDIA Isaac Sim \cite{isaac_sim} and MuJoCo \cite{todorov2012}. However, in the past years Gazebo  \cite{gazebo} has emerged as a widely used tool \cite{sarabakha2023} due to its easy integration into the ROS middleware \cite{quigley09}, a framework to exchange data in distributed networks running on different machines. It enables easy realization of multi-robot systems which can be seamlessly deployed in both simulation and on real hardware. 

In particular, in the field of aerial robotics several frameworks are available for the simulation of multi-rotor vehicles, e.g. Microsoft AirSim \cite{airsim}, a drone simulator which is realized as a plugin for Unreal Engine 4 (UE4). This enables running simulations with high graphic quality as is often required for AI-based applications.

Widely used multi-rotor drone simulators are PX4-SITL \cite{px4_sitl} and RotorS \cite{furrer2016}. Both rely on Gazebo for physics modelling and visualization thus highlighting the importance Gazebo holds within the robotic community. Whereas the former is build around the PX4 Autopilot ecosystem providing software and hardware components for autonomous control of unmanned aerial vehicles (UAVs), the latter focusses on simulation and control of multi-rotors as well as a variety of sensors.\\ 
Sphinx \cite{parrot_sphinx} is a simulator specifically designed for the simulation of the Anafi drone series distributed by Parrot Drone. Whereas both PX4-SITL and RotorS are fully open source simulators, the Anafi drone firmware and Sphinx simulator are proprietary, although customizable to a certain degree. The big advantage in using Sphinx is that it provides a testbed for  scenarios involving the high-end  drones of the Anafi series. Interfacing the drone firmware and thus retrieving the state of the drone is possible using the Olympe API \cite{parrot_olympe} for which \textit{anafi_ros} \cite{sarabakha2023} is a ROS-wrapper. This enables easy integration into ROS frameworks, however, integrating the Anafi drones into arbitrary Gazebo-based environments is not possible. \textit{anafi_autonomy} \cite{anafi_autonomy} is build on top of \textit{anafi_ros} and implements among mission planning features also a velocity controller. In order to  also provide a position and thus a tracking controller, we not only implement our own PID-based tracking framework but also an MPC which builds on top of \cite{tallamraju2018}. In this work, an efficient formulation of a MPC for tracking tasks featuring obstacle avoidance with high execution rate has been introduced. We  augment the dynamic model of the MPC with additional cummulative error states to minimize the constant tracking error while leaving out unnecessary components such as obstacle avoidance. For this task, we rely on higher-level trajectory planning methods that send waypoints to the MPC. 
Error states in model predictive control are a concept used, for example, in legged robots \cite{teng2022}, where the tracking error is defined on a Lie matrix group and linearised in the Lie algebra.  \cite{wang2023} builds on this work and extends it for the control of a quadcopter.  However, the authors point out shortcomings in the computational speed of their method. The benefit our MPC implementation is that it has even less computational cost than our customized PID controller framework.

\section{Methodology}\label{sec:Methodology} \label{sec:methodology}
\subsection{Sphinx Simulator}
 Sphinx' physics back-end is based on a customized version of Gazebo 11 and the visualization front-end on a customized version of the graphic engine Unreal Engine 4.26  \cite{parrot_unreal_engine}. However, one major drawback of Sphinx is that it cannot simulate more than one drone instance at once \cite{sphinx_one_drone_only}.

Accurate simulation of the drone in Sphinx is achieved in terms of i) realistic flight physics by means of the customized Gazebo-based back-end and ii) realistic drone handling via simulated firmware which includes the flight control, navigation and hardware management stack of the real drone into the simulation. During runtime, the firmware and thus a variety of internal drone parameters can be controlled using a Python-based API called Olympe \cite{parrot_olympe}, for which a ROS wrapper has been introduced in \textit{anafi_ros} \cite{sarabakha2023}. Furthermore, simulated drone telemetry data can be retrieved using a provided website interface or a CLI tool. 

Static actors, i.e. static meshes, can be placed in the Sphinx environment in customizable poses. Furthermore, Parrot provides a set of ready-to-use dynamic actors such as cars and pedestrians for which trajectories can be defined before runtime  \cite{parrot_sphinx_populate_with_actors} using configuration files. User generated actors with customized appearance and motion characteristics can be realized in the modified UE4 editor provided by Parrot Drone. For this purpose  Blueprint functions can be used, a graphical programming approach resembling in its workflow MATLAB / Simulink. However, as outlined in Section \ref{sec:Introduction}, state information of  actors in the scene can neither be updated nor retrieved from outside Sphinx.
Technically, numerous plugins exist for updating actor states from outside UE4-based applications, e.g. also via ROS. Retrieving state information from inside the application could also be realized using C++ code which generally can be included in the UE4 editor.  However, Parrot Drone has blocked the usage of plugins \cite{parrot_unreal_engine_no_plugins}  and custom C++ code \cite{parrot_unreal_engine_no_cpp}, necessitating other  approaches for multi-robot scenarios.

\subsection{Connection between Sphinx and Gazebo}
\begin{figure}[thpb]
      \centering
		\includegraphics[scale=0.31]{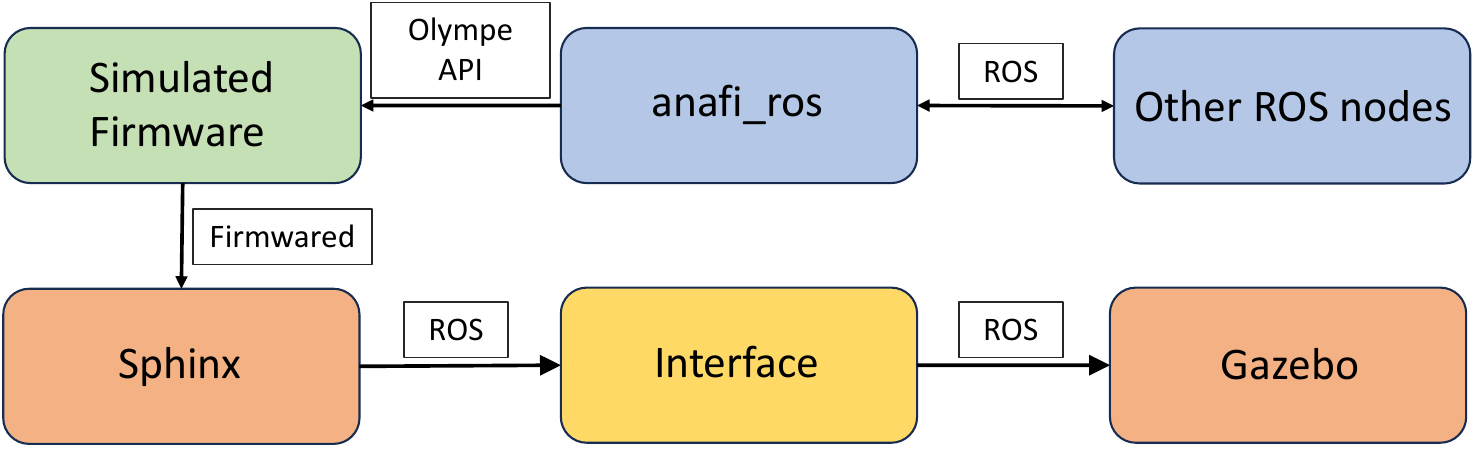}
\caption{Overview of the dataflow and connection between the components enabling the interface between Sphinx and Gazebo. Simulators are colored orange, components relying on ROS are colored blue. The simulated firmware is a stand alone module provided by Parrot Drone and colored green. Our customized set of ROS nodes providing the exchange of data between the simulators is colored yellow. The boxes next to the arrows indicate the tool managing the communication with the respective components. }
\label{fig:overview}
 \end{figure}

We propose a framework that uses ROS as central middleware to exchange data between the framework's different components.  These components not only allow to send commands to the drone in the Sphinx simulator but also to retrieve ground truth data of the drone state as well as the drone's state estimate. We use this information to set up a `mirrored' drone instance in Gazebo. This enables us  to leverage the flight behaviour of the Anafi drone simulated by Sphinx in an Gazebo-based, multi-robot environment. Figure \ref{fig:overview} illustrates the components necessary to realize the connection between Sphinx and Gazebo. 

The \textit{Sphinx} component is an instance of the Sphinx simulator running an empty world in which the Anafi drone is spawned. The drone parameters can be accessed via the simulated firmware implemented as a daemon called Firmwared, as is indicated by the \textit{Simulated Firmware} block. Parrot Drone provides a Python-based API called Olympe to communicate with the firmware. The component \textit{anafi_ros} \cite{sarabakha2023} constitutes a ROS wrapper for Olympe which allows sending and retrieving drone information via ROS topics, a necessity to realize other functionalities that are implemented elsewhere in the ROS system, e.g. controllers.
For the \textit{Interface} component we developed a set of ROS nodes accessing the Sphinx simulator via a CLI tool provided by Parrot Drone for the purpose of retrieving ground truth data about the Anafi drone. This ground truth data comprises the following. The pose and the linear and angular velocity vectors in both, world frame (an East-North-Up coordinate system) and the body center reference frame. Furthermore, the linear and angular acceleration is available in the body center reference frame only. The gimbal position and orientation and thus the pose of the camera are provided in world frame.

This information is used by the ROS nodes in the \textit{Interface} component to set up two  objects in Gazebo. The first is a robot model consisting only of a mesh  of the Anafi drone for visualization of its pose as retrieved from Sphinx.  The second is a camera plugin whose position and orientation is governed by the gimbal of the Anafi drone in Sphinx. Common to both objects is that the influence of gravity is deactivated. This avoids a jittering behaviour in the model's position which is induced by the influence of gravity in between updates of the pose. Furthermore, collision capabilities are disabled. This is due to the missing possibility to feed collision data back into Sphinx to then have an effect on the flight behaviour of the simulated drone.

A critical issue when  running two interfaced simulations in parallel is timing. Both simulations need to run with the same speed as to prevent data loss during exchanging data. Since it is not possible to update Sphinx and Gazebo on a step-wise basis, we address this issue by reducing the real time factor (RTF) to a value which can be reached by both simulations simultaneously.

\subsection{Controller Design}
We implement two types of controllers to address the problem of the Parrot Drone-provided PID-based waypoint controller showing the undesired start-stop behaviour. The first controller is PID-based using cascaded and single loop controller structures, the second is a MPC. Both control frameworks are introduced in the following. Common to both controllers is that they take  a stream of waypoints as input and generate as output setpoint values for the roll and pitch angle as well as for the yaw rate and the vertical velocity which are then sent to the Anafi \textit{Firmware} component via the \textit{anafi_ros} wrapper.
A waypoint is hereby defined as a tuple of values
\begin{equation}
WP = \left\lbrace x_{ref},y_{ref},z_{ref},v_{x,ref},v_{y,ref},v_{z,ref},\psi_{world,ref}\right\rbrace,
\end{equation} 
where $x_{ref},y_{ref},z_{ref}$ denote the desired positions, $v_{x,ref},v_{y,ref},v_{z,ref}$ the desired velocity and $\psi_{world,ref}$ the desired yaw angle in the world frame, an east-north-up (ENU) coordinate system.
\subsection{PID Controller Framework} \label{sec:pid_controller_framework}
\begin{figure}[thpb]
      \centering
		\includegraphics[scale=1]{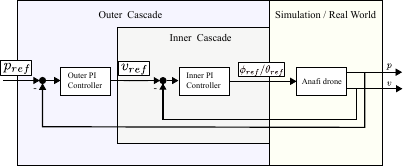}
\caption{Cascaded PID controller structure for the longitudinal and lateral motion. }
\label{fig:cascaded_PI_controller}
 \end{figure}

Under the assumptions of i) linearization around hover flight, ii) small angles, iii) a rigid, symmetric vehicle body and iv) no aerodynamic effects \cite{wang2016}, the axes of motion are decoupled  allowing to control the longitudinal, lateral, vertical and yaw movement independently using four different controllers, each controlling one direction of motion. Against this background, the first controller framework we propose consists of a set of PID controllers, in a single loop structure for the vertical velocity and yaw angle and in a cascaded structure for the longitudinal and lateral motion, the latter structure illustrated by Figure \ref{fig:cascaded_PI_controller}. In this figure, the reference position $p_{ref}$  denotes the position on the  coordinate axes associated with  the respective controller but transformed into the stability axes. The stability axes are a separate coordinate system moving along with the drone, accordingly its center resides in the drone's center of gravity. The x-axis resides in the symmetry plane of the drone in longitudinal direction and forms along with the y-axis a horizontal plane. The vertical axis points upwards, completing the right-hand system. In the stability frame the pitch angle $\theta_{ref}$ and the roll angle $\phi_{ref}$ are defined, where the pitch angle controls the longitudinal and the roll angle the lateral motion. The desired yaw angle is set via the yaw rate and the vertical motion via the vertical velocity.

\subsection{Quadratic Model Predictive Controller}\label{sec:mpc}
In order to reduce the tracking error present for PID based tracking controllers, we focus on the reduction of the offset between the moving target waypoint and the drone during the design of the MPC by introducing cumulative error states. In this regard, using a quadratic formulation of the MPC is beneficial as convex quadratic optimization problems are fast to solve and computationally cheap \cite{boyd2004}.

We use the implementation of a constrained linear-quadratic MPC using the solver OSQP \cite{osqp} and define the following finite-horizon optimal control problem to be solved at each time step
\begin{equation}
\begin{split}
\begin{array}{ll}
\mathbf{x}^*_0,\mathbf{x}^*_1, \ldots, \mathbf{x}^*_{N}& ,\mathbf{u}^*_0,\mathbf{u}^*_1, \ldots, \mathbf{u}^*_{N} \\
 = \underset{\mathbf{u}_0,\mathbf{u}_1, \ldots, \mathbf{u}_{N}}{ \mbox{arg min} }  & (\mathbf{x}_N-\mathbf{z}_{N})^T \mathbf{Q}_N (\mathbf{x}_N-\mathbf{z}_{N}) +\\
  					 & \sum_{k=0}^{N-1} (\mathbf{x}_k-\mathbf{z}_{k})^T \mathbf{Q} (\mathbf{x}_k-\mathbf{z}_{k}) + \mathbf{u}_k^T \mathbf{R} \mathbf{u}_k \\
  \mbox{subject to} & \mathbf{x}_{k+1} = \mathbf{A} \mathbf{x}_k + \mathbf{B} \mathbf{u}_k \\
  					 & \mathbf{z}_{k+1} = \mathbf{A}_r \mathbf{z}_{k}  \\
                    & \mathbf{x}_{\rm min} \le \mathbf{x}_k  \le \mathbf{x}_{\rm max} \\
                    & \mathbf{u}_{\rm min} \le \mathbf{u}_k  \le \mathbf{u}_{\rm max} \\
                    & \mathbf{x}_0 = \bar{\mathbf{x}}\\
                    & \mathbf{z}_{0} =\bar{\mathbf{z}},
\end{array}
\end{split}\label{eq:MPC}
\end{equation} 
where $\mathbf{z}_{k}\in\mathbb{R}^{n_x}$ denotes the reference state vector that shall be tracked by the state vector  $\mathbf{x}_k\in\mathbb{R}^{n_x}$, with $n_x\in\mathbb{N}$ states each. The input vector to the dynamic model   $\mathbf{x}_{k+1} = \mathbf{A} \mathbf{x}_k + \mathbf{B} \mathbf{u}_k $ is  $\mathbf{u}_k\in\mathbb{R}^{n_u}$ with $n_u\in\mathbb{N}$ inputs.   $\mathbf{x}_{\rm min}$,  $\mathbf{x}_{\rm max}$, $\mathbf{u}_{\rm min}$ and $\mathbf{u}_{\rm max}$ denote the state and input limits. The initial states  $\mathbf{x}_0 = \bar{\mathbf{x}}$ and $\mathbf{z}_{0}= \bar{\mathbf{z}}$ are updated repeatedly for each execution of the optimization problem. In each execution cycle the MPC computes an optimal trajectory for the controller input, $\mathbf{u}^*_0,\mathbf{u}^*_1,\mathbf{u}^*_2 \ldots \mathbf{u}^*_{N} $ which lead to the optimal state trajectory  $\mathbf{x}^*_0,\mathbf{x}^*_1,\mathbf{x}^*_2 \ldots \mathbf{x}^*_{N} $ that best tracks the reference states $\mathbf{z}_k$. 
We define the reference state vector $\mathbf{z}_k$ and state vector $\mathbf{x}_k$ as follows.
\begin{dmath}
\mathbf{z}_k = \left[x_{ref,k},  x_{ref,k},  y_{ref,k},  y_{ref,k},  z_{ref,k},  z_{ref,k},...\\ v_{x,ref,k},v_{x,ref,k}, v_{y,ref,k}, v_{y,ref,k},  v_{z,ref,k},v_{z,ref,k},...  \\x_{cum,ref,k},  y_{cum,ref,k},  z_{cum,ref,k},...\\ v_{x,cum,ref,k},  v_{y,cum,ref,k}, v_{z,cum,ref,k}\right]^T,
\end{dmath}
\begin{dmath}
\mathbf{x}_k = \left[x_{ref,k},  x_k,  y_{ref,k},  y_k,  z_{ref,k},  z_k,...\\ v_{x,ref,k},v_{x,k}, v_{y,ref,k}, v_{y,k},  v_{z,ref,k},v_{z,k},...  \\x_{cum,k},  y_{cum,k},  z_{cum,k},...\\ v_{x,cum,k},  v_{y,cum,k}, v_{z,cum,k}\right]^T,
\end{dmath}
where $x_{ref,k}$, $y_{ref,k}$, $z_{ref,k}$, $v_{x,ref,k}$, $v_{y,ref,k}$, $v_{z,ref,k}$ denote the waypoint's position and velocity coordinates, $x_k$, $y_k$, $z_k$, $v_{x,k}$, $v_{y,k}$, $v_{z,k}$ the position and velocity coordinates of the drone, and $x_{cum,ref,k}$, $ y_{cum,ref,k}$, $ z_{cum,ref,k}$ as well as $v_{x,cum,ref,k}$, $ v_{y,cum,ref,k}$, $v_{z,cum,ref,k}$ denote the reference values for the cumulative errors $x_{cum,k}$, $y_{cum,k}$, $z_{cum,k}$ and $v_{x,cum,k}$, $v_{y,cum,k}$, $v_{z,cum,k}$, respectively, each at the prediction step $k \in \left\lbrace 0,1,2,\ldots,N \right\rbrace$.
The reason why the reference values are part of the state vector $\mathbf{x}$  is that they need to be included in the formulation of the dynamic model as to be able to build the cumulative error necessary for reducing the tracking offset. This is also why the reference values appear twice in the reference vector $\mathbf{z}_k$.

The dynamics model $\mathbf{x}_{k+1} = \mathbf{A} \mathbf{x}_k + \mathbf{B} \mathbf{u}_k $ of the MPC is defined by the matrices $\mathbf{A},\mathbf{B}$ as follows.
\begin{equation}
\mathbf{A} =
\begin{bmatrix}
    \mathbf{I}_6 & \Delta t \mathbf{I}_6 & \mathbf{0}_{6 \times 6} \\
    \mathbf{0}_{6 \times 6} & \mathbf{I}_6 & \mathbf{0}_{6 \times 6} \\
	 &	\begin{bmatrix}
		1 & -1 & 0 & \cdots & 0 \\
		0 & 1 & -1 & \cdots & 0 \\
		0 & 0 & 1 & \cdots & 0 \\
		0 & 0 & 0 & \ddots & 0 \\
		\vdots & \vdots & \vdots & \ddots & 0 \\
		0 & 0 & 0 & \cdots & -1 
	\end{bmatrix}_{6\times 12} & \mathbf{I}_{6 \times 6}
\end{bmatrix}
\end{equation}

The upper part consisting of the first two rows of matrices defines a simple mass point model. The lower $6\times12$ matrix computes the difference in position and velocity between the waypoint and the drone. This offset is added to the existing cumulative error as indicated by the matrix $\mathbf{I}_{6\times6}$ in the lower right corner. 
The input vector to the dynamics model $\mathbf{u}$  is defined as
\begin{dmath}
\mathbf{u} = \left[a_x,a_y,a_z\right]^T.
\end{dmath}
The input matrix $\mathbf{B} = \left[ \mathbf{b}_1,\mathbf{b}_2, \mathbf{b}_3 \right]$  specifies how the control inputs $\mathbf{u}$ influence the states associated with the drone's position and velocity.
\begin{align}
\mathbf{b}_1 & = \left[\mathbf{0}_{1\times1},\Delta t^2/2 , \mathbf{0}_{1\times5},\Delta t ,\mathbf{0}_{1\times10} \right]^T \notag   \\ 
\mathbf{b}_2 & = \left[\mathbf{0}_{1\times3},\Delta t^2/2 , \mathbf{0}_{1\times5},\Delta t ,\mathbf{0}_{1\times8} \right]^T\\
\mathbf{b}_3 & = \left[\mathbf{0}_{1\times5},\Delta t^2/2 , \mathbf{0}_{1\times5},\Delta t ,\mathbf{0}_{1\times6} \right]^T  \notag
\end{align}

The matrix $\mathbf{A}_{r}$ defines a simple constant velocity model which is used to update the reference state $\mathbf{z}_{k}$ over the prediction horizon. In tracking scenarios, it is commonly assumed that each waypoint $WP$ is associated with a velocity vector, which defines the expected motion of the waypoint over time.
\begin{equation}
\mathbf{A}_{r} =
\begin{bmatrix}
\mathbf{I}_{6\times6} & \begin{bmatrix} \Delta t & 0 \\ \Delta t & 0\end{bmatrix} \otimes \mathbf{I}_{3\times3} & \mathbf{0}_{6\times6} \\ 
\mathbf{0}_{6\times6} & \begin{bmatrix} 1 & 0 \\ 1 & 0\end{bmatrix} \otimes \mathbf{I}_{3\times3} & \mathbf{0}_{6\times6} \\ 
&\mathbf{0}_{6\times18}&\\
\end{bmatrix}
\end{equation}
Careful testing has revealed that best tracking capabilities are achieved when $a_{x}, a_{y}$ taken from $\mathbf{u}_0^*$ (see \eqref{eq:MPC})  are published to the drone. For the longitudinal and lateral motion, these  accelerations are translated via the following equations into the required pitch angle $\theta$ and the roll angle $\phi$.

\begin{align}
\theta &= \sin^{-1}(a_{x} / g)\\
\phi &= \sin^{-1}(a_{y} / g)
\end{align}
For the required vertical velocity to set the altitude of the drone the value $v_{z}$ taken from $\mathbf{x}_5^*$ (see \eqref{eq:MPC}) has proven most successful. The yaw angle is not controlled by the MPC but by the same single loop PID controller used in the framework described in Section \ref{sec:pid_controller_framework}.

\section{Implementation}\label{sec:implementation}
\subsection{General}
We set up the experiments to demonstrate
\begin{itemize}
\item  that our framework allows the integration of the Anafi drone into a complex multi-robot simulation in Gazebo.
\item  that our MPC  tracking controller (see Section \ref{sec:mpc}) outperforms in simulation and real world experiments  our customized PID controller framework (see Section \ref{sec:pid_controller_framework}). We introduced the latter to serve as a  baseline for the performance evaluation of the MPC while addressing the shortcomings of the PID control stack which is implemented in the Anafi firmware and provided by Parrot Drone.
\end{itemize}
\subsection{Hardware}
All simulations were conducted on a desktop computer with the following specifications. Ubuntu 20, AMD Ryzen threadripper 3960x 24-core processor, NVIDIA GeForce RTX 3080, 128 GB RAM, 6TB SSD. The installed ROS version is  Noetic featuring Gazebo 11.11, ROS nodes are written in Python 3.8.10. The installed version of Sphinx is 2.15.1.

In the real-world experiments, all ROS components were run on a laptop featuring Ubuntu 20, an Intel Core i7-6600U CPU, an Intel HD Graphics 520 GPU, 32 GB RAM and a 512 GB SSD.

\subsection{Integration of the Anafi Drone into a Multi-Robot Simulation}
We choose our Gazebo-based airship simulation \cite{price2022} to include the Anafi drone as a second aerial vehicle in the simulation. The \textit{Interface} component sends data from Sphinx to Gazebo with a frequency of $100\si{hz}$. 

\subsection{Verification of the Custom PID-based Controller Framework and of the MPC}
\setlength{\tabcolsep}{4pt}

\begin{table}[]
\center
\small
\caption{Parameters set for the PID controller and the MPC. \label{tab:controller_constraints}}
\begin{tabular}{@{}lll@{}}
\toprule
Controller& Parameter           & Value \\ \midrule
MPC & $ \left\lbrace x,y,z \right\rbrace_{max/min} $ & $\pm \infty \si{m}$ \\
MPC & $ \left\lbrace x_{ref},y_{ref},z_{ref} \right\rbrace_{max/min} $ & $\pm \infty \si{m}$ \\
MPC &$ \left\lbrace v_x,v_y,v_z \right\rbrace_{max/min} $ & $\pm \left\lbrace 10,10, 2\right\rbrace \si{m/s}$ \\
MPC &$ \left\lbrace v_{ref,x},v_{ref,y},v_{ref,z} \right\rbrace_{max/min} $ & $\pm \infty \si{m/s}$ \\
MPC &$ \left\lbrace x,y,z \right\rbrace_{cum,max/min} $ & $\pm \infty \si{m}$ \\
MPC &$ \left\lbrace v_x,v_y,v_z\right\rbrace_{cum,max/min}$ & $\pm \infty \si{m/s}$ \\
MPC &$ \left\lbrace a_x,a_y,a_z \right\rbrace_{max/min} $ & $\pm 5 \si{m/s^2}$ \\
MPC &$N$ & $30$\\
MPC &$\Delta t$ & $0.1$\\
MPC &Execution/output frequency & $10\si{hz}/10\si{hz}$\\

\midrule
PID &$ \left\lbrace v_x,v_y,v_z \right\rbrace_{max/min} $ & $\pm \left\lbrace 10,10, 2\right\rbrace \si{m/s}$  \\
PID &$ v_{z,max/min}$ & $\pm 2\si{m/s}$\\
PID &$ \theta_{max/min}$ & $\pm 30.64\si{^{\circ}}$\\
PID &$ \phi_{max/min}$ & $\pm 30.64\si{^{\circ}}$\\
PID &Yaw rate & $\pm 180\si{^{\circ}/s}$\\
PID &Execution/output frequency & $\sim30\si{hz}/10\si{hz}$\\
\bottomrule
\end{tabular} 
\end{table}
Table \ref{tab:controller_constraints} summarizes the values of parameters associated with the PID controller framework or the MPC, respectively.

\setlength{\tabcolsep}{6pt}
\begin{table}[]
\center
\small
\caption{Controller gains used for the (cascaded) PID-based controller framework. \label{tab:gains}}
\begin{tabular}{@{}ccccc@{}}
\toprule
Control Loop            & $k_p$ & $k_i$  & $k_d$ &  \\ \midrule
Lon. outer loop & 0.875 & 0.0001 & 0     &  \\
Lon. inner loop & 10    & 1      & 0     &  \\
Lat. outer loop      & 0.875 & 0.001  & 0     &  \\
Lat. inner loop      & -10    & -1      & 0     &  \\
Vertical             & 1     & 0      & 0.001 &  \\
Yaw                  & 0.6   & 0.2    & 0.003 & \\
\bottomrule
\end{tabular} 
\end{table}

The PID controller baseline is implemented using the ROS package \textit{pid} \cite{ros_pid} in combination with custom interface nodes. Table \ref{tab:gains} summarizes the gain values used for the different controllers.
The weight matrices used in \eqref{eq:MPC} were carefully tuned to achieve the best possible behaviour and are defined as follows.
\begin{align}
\mathbf{Q} &= \operatorname{diag}\left( \mathbf{0}_{1\times12} ,10^{-3},10^{-3},10^{-3},\mathbf{0}_{1\times3}\right)\\
\mathbf{Q_N} &= \operatorname{diag}\left( \mathbf{0}_{1\times12} ,5,5,5,\mathbf{0}_{1\times3}\right)\\
\mathbf{R} &= \operatorname{diag}\left( 0.2,0.2,0.2\right)\\
\end{align}
We also found that weights  for only the positional cumulative errors were required to achieve good tracking performance.
The initial values of $\bar{\mathbf{z}}$ are based on the  commanded waypoints.
\begin{dmath}
\bar{\mathbf{z}}= \left[x_{ref},  x_{ref},  y_{ref},  y_{ref}, z_{ref},  z_{ref},,...\\ v_{x,ref},v_{x,ref}, v_{y,ref}, v_{y,ref},  v_{z,ref},v_{z,ref},...  \\ 0,  0,  0,0, 0, 0\right]^T,
\end{dmath}
Whereas the first 12 reference  states follow the waypoint values, the last six states are set to 0. They define the desired cumulative error for the positional and velocity states that the solver should achieve when optimizing \eqref{eq:MPC} 

\section{Results}\label{sec:experiments}
\subsection{Integration of the Anafi Drone into a Gazebo Simulation}
\begin{figure}[thpb]
      \centering
		\includegraphics[scale=0.2]{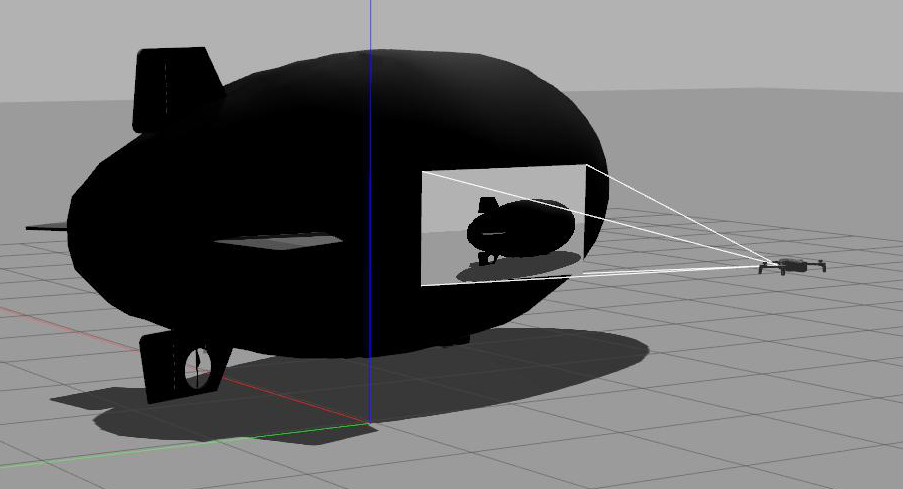}
\caption{\textit{Anafi} drone and airship set up jointly in a common Gazebo simulation environment. }
\label{fig:anafi_blimp}
 \end{figure}

Figure \ref{fig:anafi_blimp} shows the Gazebo-based airship simulation \cite{price2022} including a mirrored instance of the Anafi drone. 
Since Gazebo can have difficulties loading and rendering multiple complex meshes simultaneously, we neglect the display of rotating propellers to keep the model of the \textit{Anafi} drone as simple as possible. On our setup described in Section \ref{sec:implementation} where both simulators run on a single computer, we achieved a real time factor of $RTF = 0.6$. However,  one advantage of the ROS middleware used for exchanging data between the simulators is that it can run on distributed systems, thus providing the possibility to increase the real time factor by running different components of the system illustrated by Figure \ref{fig:overview} on different machines. As a consequence, a real time factor of $RTF = 1$ was achieved when Sphinx was installed on a separate computer with near identical specifications than the computer hosting Gazebo.
 
\subsection{Evaluation of Tracking Controllers}
\subsection*{Computational Workload}
On the computational platform used for the simulation experiments, the CPU time required by our customized PID framework was approximately 2.59 times greater than that of the MPC, indicating a substantially higher computational burden. This increased cost is likely attributable to the distributed architecture of the PID framework, wherein each control cascade is deployed as an independent node. Additional nodes are responsible for tasks such as computing relative positional information between the drone and its waypoints, as well as managing the collection and redistribution of data among the various control cascades. In contrast, the MPC is formulated as a quadratic optimization problem, which can be solved with high computational efficiency.

\subsection*{Evaluation Scenarios}
For the evaluation and comparison of our MPC and PID controller's tracking capabilities of static and moving waypoints in simulation and the real world we design four different scenarios. i) Static waypoints, ii) a straight line trajectory at different speeds, iii)  different circular trajectories residing in the $x,y$-plane with different radii and  speeds and iv) a rectilinear periodic movement in the $z$-direction only. The reason is that these scenarios provide standardized movements allowing a fair comparison between the controllers.

\subsubsection{Static Waypoints}
\begin{figure}[thpb]
      \centering
		\includegraphics[scale=0.35]{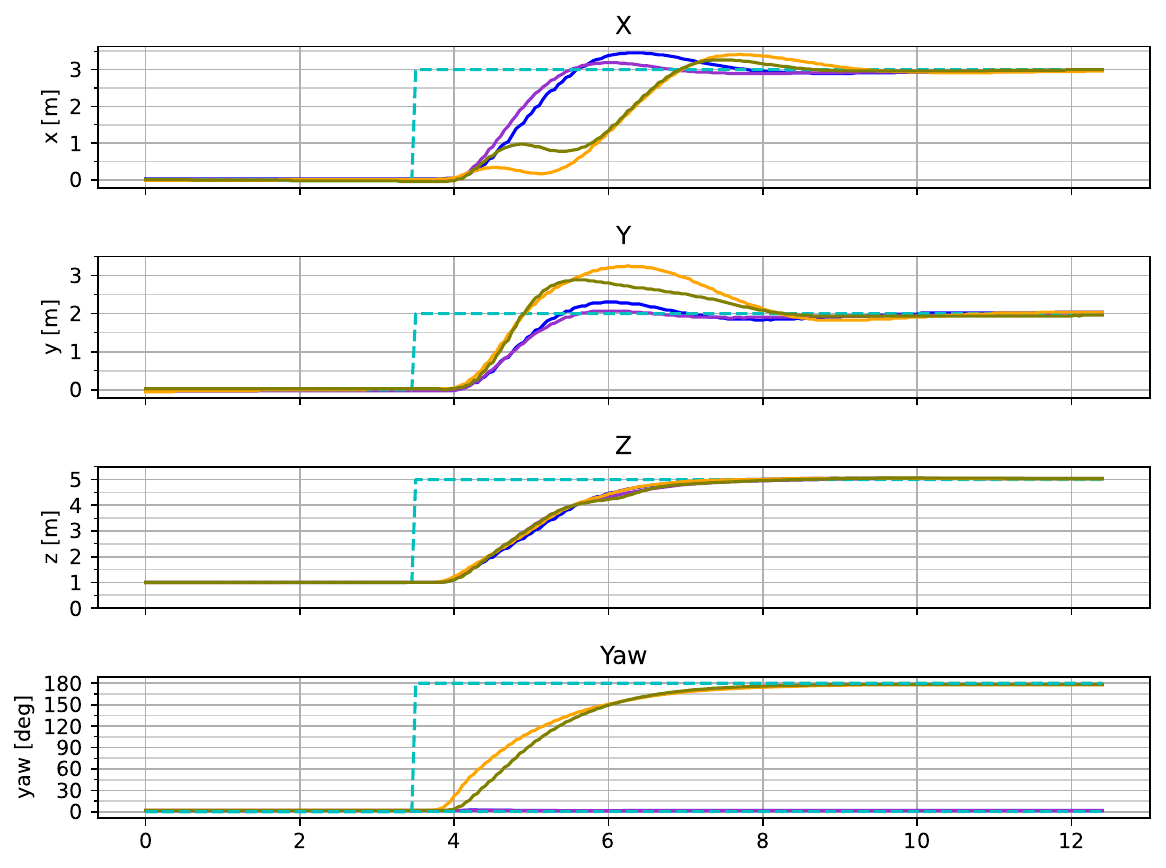}
\caption{Comparison between the performance of the PID-based controller framework and MPC in simulation. The waypoints are $WP_0=\left\lbrace 3,2,5,0,0,0,0\right\rbrace$ (PID: blue, MPC: purple) and $WP_1=\left\lbrace 3,2,5,0,0,0,180^{\circ}\right\rbrace$  (PID: orange, MPC: olive)}
\label{fig:static_waypoints_sim}
 \end{figure}

Figure \ref{fig:static_waypoints_sim} qualitatively compares in simulation the performance between our customized PID framework and the MPC for reaching static waypoints starting in hover flight. Two waypoints are examined. The difference between them is the yaw angle at which the waypoints should be reached. For the first waypoint $WP_0$ it is $\psi_{ref,world} = 0\si{rad}$, for the second waypoint $WP_1$ is $\psi_{ref,world} = \pi\si{rad}$ .
It can be seen that being a predictive controller type, the MPC shows significantly less overshoot than the PID controller framework. The large overshoot for $WP_{1}$ in the $y$-direction is due to the yaw rotation which is conducted by a separate control loop for both the cascaded PID controller framework and the MPC. This is why even the MPC can only react to the external disturbance as the yaw movement is not considered in the MPC formulation.
\begin{figure}[thpb]
      \centering
		\includegraphics[scale=0.35]{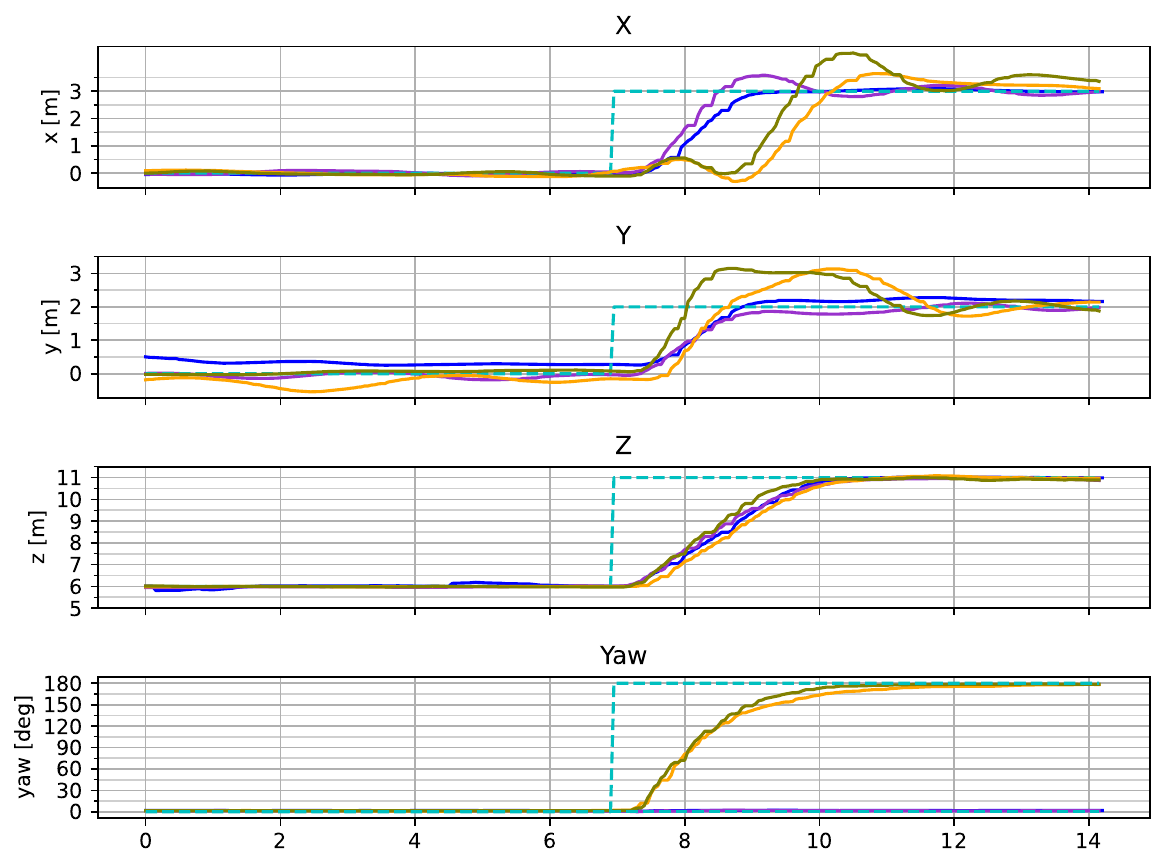}
\caption{Comparison between the performance of the PID-based controller framework and MPC in the real world. The waypoints are $WP_0=\left\lbrace 3,2,11,0,0,0,0\right\rbrace$ (PID: blue, MPC: purple) and $WP_1=\left\lbrace 3,2,11,0,0,0,180^{\circ}\right\rbrace$  (PID: orange, MPC: olive)}
\label{fig:static_waypoints_real}
 \end{figure}
 Figure \ref{fig:static_waypoints_real} illustrates the same scenario but in  the real world experiments. It can be seen that this time the MPC is not beneficial in comparison to the PID controller framework. However, this could be attributed to the windy conditions ($v_{wind} = 4\si{m/s}$ with considerably higher wind gusts possibly present when testing the MPC). Another possible explanation are the cumulated error states in combination with real-world effects  not modelled in Sphinx that could lead to a slightly more pronounced overshoot of the waypoint than with the  PID controller framework. However, their advantage in dynamic waypoint scenarios outweighs this aspect. 
\subsubsection{Straight Line Trajectory}
The purpose of testing the tracking performance of the controllers for a target moving on a straight line is to evaluate the tracking error $e$. Against this background, moving targets were evaluated for different speeds as is summarized in Table \ref{tab:straight_line_evaluation}.
\begin{table}[]
\center
\tiny
\caption{ Mean tracking error in meters between drone and waypoint moving on a straight line for the different control frameworks and different target speeds in simulation.\label{tab:straight_line_evaluation}}
\begin{tabular}{@{}cccccc@{}}
\toprule
Contr.$\downarrow$ / Speed$\rightarrow$ & $1\si{m/s}$ &  $2\si{m/s}$  &  $3\si{m/s}$ &  $4\si{m/s}$ & $5\si{m/s}$\\ 
\midrule 
        MPC &   0.16   &  0.36   &   0.59    & 0.82  & 1.41  \\
        PID &   1.32   &   2.67   &  4.07   &  5.32  &7.00\\
\bottomrule\\
\end{tabular}
\end{table}
It is clear that the MPC achieves a significantly better tracking accuracy, with the tracking error $e$ staying in the range of  $1\si{m}$ as compared to the  $6\si{m}$ of the PID.  Real world experiments were not conducted for this scenario due to safety issues with the restricted flight space and concerns about losing the Anafi drone out of sight. 
\subsubsection{Circular Trajectory} In this scenario the target moves in a circle at a fixed altitude with a constant yaw angle $\psi_{ref,world}=0\si{rad}$. Different radii and moving speeds were tested. This scenario was chosen to evaluate the tracking error $e$ under more complex conditions. The results are summarized in Table \ref{tab:circular_trajectory_evaluation_sim} for the experiments conducted in simulation and  the real world.

\setlength{\tabcolsep}{4pt}

\begin{table}[]
\center
\tiny
\caption{Maximum tracking error $e_{max}$, mean tracking error $e_{\mu}$, and standard deviation $e_{\sigma}$ of the tracking error for different circular trajectories defined by the radius $r$ and the velocity $v$ of the tracked target for flights conducted in simulation and the real world. \label{tab:circular_trajectory_evaluation_sim}}
\begin{tabular}{c cc ccc ccc}
\toprule
{Controller} & \multicolumn{2}{c}{{Circle}} & \multicolumn{3}{c}{{Simulation}} & \multicolumn{3}{c}{{Real World}} \\ 
\cmidrule(lr){1-1} \cmidrule(lr){2-3} \cmidrule(lr){4-6} \cmidrule(lr){7-9}
Name & $r [\si{m}]$ & $v [\si{m/s}]$ & $e_{max} [\si{m}]$ & $e_{\mu} [\si{m}]$ & $e_{\sigma} [\si{m}]$ & $e_{max} [\si{m}]$ & $e_{\mu} [\si{m}]$ & $e_{\sigma} [\si{m}]$ \\ 
\midrule 
MPC & 1 & 1 & 0.73 & 0.66 & 0.03 & 0.72 & 0.57 & 0.05 \\ 
PID & 1 & 1 & 1.38 & 1.32 & 0.03 & 1.69 & 1.29 & 0.21 \\ 
\midrule 
MPC & 2 & 1 & 0.47 & 0.39 & 0.03 & 0.61 & 0.44 & 0.07 \\ 
PID & 2 & 1 & 1.37 & 1.25 & 0.05 & 1.54 & 1.33 & 0.07 \\ 
\midrule 
MPC & 2 & 2 & 1.63 & 1.44 & 0.10 & 1.39 & 1.11 & 0.12 \\ 
PID & 2 & 2 & 2.70 & 2.57 & 0.05 & 2.70 & 2.36 & 0.09 \\ 
\midrule 
MPC & 3 & 1 & 0.43 & 0.33 & 0.03 & 0.59 & 0.35 & 0.07 \\ 
PID & 3 & 1 & 1.49 & 1.33 & 0.06 & 1.56 & 1.36 & 0.07 \\ 
\midrule 
MPC & 3 & 2 & 1.37 & 1.19 & 0.07 & 1.20 & 0.94 & 0.09 \\ 
PID & 3 & 2 & 2.74 & 2.57 & 0.09 & 3.20 & 2.54 & 0.27 \\ 
\midrule 
MPC & 3 & 3 & 2.66 & 2.43 & 0.07 & 1.98 & 1.54 & 0.14 \\ 
PID & 3 & 3 & 4.42 & 4.26 & 0.07 & 4.23 & 3.70 & 0.23 \\ 
\bottomrule 
\end{tabular} 
\end{table}

The MPC shows better performance than the PID across all scenarios, although at higher velocities the advantage of the MPC diminishes. The reason is that in these scenarios the  control commands produced by the MPC are bound by the maximum attitude angles of the drone thus limiting its manueverability. 
The real world experiments confirm the simulation results, although they were conducted under windy conditions subjecting the drone to heavy disturbances. However, the standard deviation of the tracking error $e_{\sigma}$ of the MPC is considerably lower than the one of the PID controller framework thus indicating that the tracking is managed in an evenly way.

\subsubsection{Vertical Rectilinear Periodic Trajectory}
The tracking capability in vertical direction is verified for a target conducting a rectilinear periodic trajectory (RPT) in vertical direction only. Different radii and velocities of the RPT are tested. 
Table \ref{tab:rpm_trajectory_evaluation_sim} summarizes the results for the experiments conducted in simulation and the real world.
\begin{table}[]
\center
\tiny
\caption{Maximum tracking error $e_{max}$, mean tracking error $e_{\mu}$, and standard deviation $e_{\sigma}$ of the tracking error for different rectilinear periodic trajectories (RPT) defined by the radius $r$ and the velocity $v$ of the tracked target for flights conducted in simulation and the real world. \label{tab:rpm_trajectory_evaluation_sim}}
\begin{tabular}{c cc ccc ccc}
\toprule
{Controller} & \multicolumn{2}{c}{RPT} & \multicolumn{3}{c}{Simulation} & \multicolumn{3}{c}{Real World} \\ 
\cmidrule(lr){1-1} \cmidrule(lr){2-3} \cmidrule(lr){4-6} \cmidrule(lr){7-9}
Name & $r [\si{m}]$ & $v [\si{m/s}]$ & $e_{max} [\si{m}]$ & $e_{\mu} [\si{m}]$ & $e_{\sigma} [\si{m}]$ & $e_{max} [\si{m}]$ & $e_{\mu} [\si{m}]$ & $e_{\sigma} [\si{m}]$ \\ 
\midrule 
MPC & 1 & 1 & 0.36 & 0.21 & 0.10 & 0.64 & 0.31 & 0.11 \\ 
PID & 1 & 1 & 0.96 & 0.60 & 0.29 & 1.08 & 0.68 & 0.26 \\ 
\midrule 
MPC & 2 & 1 & 0.25 & 0.13 & 0.06 & 0.37 & 0.18 & 0.07 \\ 
PID & 2 & 1 & 1.07 & 0.65 & 0.32 & 1.12 & 0.65 & 0.30 \\ 
\midrule 
MPC & 2 & 2 & 1.08 & 0.51 & 0.28 & 1.27 & 0.72 & 0.31 \\ 
PID & 2 & 2 & 1.91 & 1.20 & 0.57 & 2.10 & 1.31 & 0.54 \\ 
\bottomrule
\end{tabular} 
\end{table}

Similar to the circular trajectories, the results show that the prediction capability of the MPC  significantly improves the tracking performance compared to the PID controller framework. The reason why the tracking error is even more reduced by the MPC compared to the PID-based controller framework and also compared to the circular trajectory is that during the RPT the drone only moves in the vertical direction instead of the  longitudinal and lateral direction simultaneously, thus constituting a simpler motion. Both controllers perform slightly worse in the real world than in simulation although for the MPC the loss in performance is less pronounced. However, similar to the circular trajectories, the standard deviation of the tracking error $e_{\sigma}$ is considerably lower than the one of the PID controller framework across all scenarios indicating that the tracking performance of the MPC is maintaining a  more consistent quality. 

The question of which tracking accuracy is required strongly depends on the mission. Whereas landing on a moving platform may only allow a tolerance within the range of less than $0.5\si{m}$, high altitude surveillance tasks might accept a tracking error of several meters. However, the detailed experiment evaluation presented above allows to compensate the remaining tracking error by shifting the waypoint in the direction of motion based on the mean tracking error associated with the current speed of the Anafi drone.

\section{Conclusion} \label{sec:conclusion}

This work integrates the Parrot Sphinx simulator with Gazebo-based multi-robot frameworks and addresses limitations of the Anafi drone's default PID controller. We introduce two alternative controllers for tracking tasks: an improved PID controller with limited tracking accuracy, and a Model Predictive Controller (MPC) that reduces tracking error using a modified dynamics model with cumulative error states. Integration is demonstrated in our airship simulation \cite{price2022}, and both controllers are evaluated in simulation and real-world scenarios. We aim for this work to support the broader adoption of the Anafi drone in scientific research applications. Future work will address the current limitation of supporting only a single Anafi drone instance within a Gazebo simulation environment.

 \bibliographystyle{IEEEtran}
 \bibliography{./bib/bibliography}

\end{document}